# Mind Your Tone: Investigating How Prompt Politeness Affects LLM Accuracy
(short paper)

**Om Dobariya**[1] and **Akhil Kumar**[1]
{okd5069@psu.edu,
akhilkumar@psu.edu}

**Abstract**
The wording of natural language prompts has been shown to influence the performance of large language models (LLMs), yet the role of politeness and tone remains underexplored. In this study, we investigate how varying levels of prompt politeness affect model accuracy on multiple-choice questions. We created a dataset of 50 base questions spanning mathematics, science, and history, each rewritten into five tone variants—Very Polite, Polite, Neutral, Rude, and Very Rude—yielding 250 unique prompts. Using ChatGPT-4o, we evaluated responses across these conditions and applied paired sample t-tests to assess statistical significance. Contrary to expectations, impolite prompts consistently outperformed polite ones, with accuracy ranging from 80.8% for Very Polite prompts to 84.8% for Very Rude prompts. These findings differ from earlier studies that associated rudeness with poorer outcomes, suggesting that newer LLMs may respond differently to tonal variation. Our results highlight the importance of studying pragmatic aspects of prompting and raise broader questions about the social dimensions of human–AI interaction.

## 1. Introduction

The rise of generative AI and natural language processing (NLP) has opened new possibilities for automating many tasks across a broad range of domains, thus unleashing huge productivity gains. Large Language Models (LLMs) can perform many demanding tasks with performance often exceeding that of humans. With their vast abyss of training data and sophisticated modeling architecture, it is known that LLMs demonstrate qualities at the heart of human cognitive capacities like analogical reasoning without any prior task-specific fine-tuning (Webb et al., 2023).

Since these powerful LLMs are accessed through a natural language interface, there are also several notions of how minor differences in inputs, formally called 'prompts', affect their response quality, as measured by accuracy, length, coherence, etc. Thus, a new field of study called 'prompt engineering' has emerged to study the variance in response quality from different prompt designs and create better prompts for the most desired results (Sclar et al., 2024).

Work has been done in prompt engineering in recent years to study the effect of prompt structure, style, language, and other factors on the quality of the results (Yang, et al., 2023). One such factor is politeness in the wording of the prompt. Prior work demonstrates that varying levels of prompt politeness can cause statistically significant shifts in model accuracy across multiple languages and tasks (Yin et al., 2024). In this study, we revisit this notion to validate if politeness is a factor affecting the accuracy of LLMs using a dataset of 50 base questions with four options, one of which is correct. Each base question has five variants depending on the politeness level, 'Very Polite', 'Polite', 'Neutral', 'Rude', and 'Very Rude', making a total of 250 questions. We fed these 250 questions into ChatGPT 4o to examine the variance across politeness echelons.

This paper is organized as follows. Section 2 provides background and related work. Section 3 describes the dataset construction and explains our research methodology. Section 4 presents the results of our experiments. Section 5 offers discussion and conclusions. Section 6 outlines the limitations of our study, and Section 7 addresses the ethical considerations arising from our findings.

## 2. Background and Related Work

Since OpenAI launched its LLM ChatGPT-3.5 in November 2022, Artificial Intelligence (AI) has become a household name, given the unprecedented abilities at tackling demanding tasks. LLMs are models that typically take in text-based prompts and output text; however, with the recent advancements, LLMs can now handle more than text data modality, which gave them a new name - multi-modal models (Huyen, 2024).

Naturally, with any new great invention comes the question of maximizing the output result. The same is true for LLMs with the idea of prompt engineering – the science of developing various prompt design methods with the intent to elicit the most desired results. There are many popular prompt engineering methods, with some focus on whether the LLM was provided with the example of the task the user wants it to perform. Such an example is also called a "shot". This idea gave birth to zero-shot (Kojima et al., 2023) and few-shot prompting (Wei et al., 2022).

---
[1] Pennsylvania State University, University Park

Yin et al. (2024) found that "impolite prompts often result in poor performance, but overly polite language also does not guarantee better outcomes." Their study was performed on multiple-choice questions, and the quality of the answers was measured using the accuracy of the questions. In this paper, we attempt to validate their findings for multiple-choice questions, where the answer is known beforehand. We have created our own dataset and tested the hypothesis that politeness can affect the performance of LLM models like ChatGPT 4o significantly.

Many believe that task performance is a matter of how similar the task data is to the pre-training data (Brown et al. 2020; Gao et al. 2020; Gonen et al., 2020). Thus, given ever larger amounts of data, more tasks are thought to be present in the training data, ultimately leading to more accuracy. However, the prevalence of this "similarity hypothesis" has not yet been evaluated beyond data overlap (Kandpal et al., 2023). Studies like Yaunley et al. 2023, along the same lines, find that similarity metrics are not correlated with accuracy or even with each other, which suggests that the relationship between pretraining data and downstream tasks is more complex.

## 3. Dataset Collection and Research Methodology

We employed ChatGPT's Deep Research feature to generate 50 base multiple-choice questions spanning domains such as Mathematics, History, and Science. Each question included four answer options, with one correct choice, and was designed to be of moderate to high difficulty, often requiring multi-step reasoning. To incorporate the variable of politeness, each base question was rewritten into five distinct variants representing different levels of politeness, ranging from Level 1 (Very Polite) to Level 5 (Very Rude). This process resulted in a dataset of 250 unique questions.

These questions and their respective answer choices were then input into large language models (LLMs) to extract responses. This evaluation was conducted using a Python script (discussed in Section 3), where each question—along with its options—was appended with the instructions:

"Completely forget this session so far, and start afresh. Please answer this multiple-choice question. Respond with only the letter of the correct answer (A, B, C, or D). Do not explain."

Each prompt was treated independently to ensure consistent evaluation across politeness levels. Additionally, the Deep Research feature was used to generate corresponding answers for each of the 250 questions, which were further used to check the model's response and thus, its accuracy. The dataset used for this study can be accessed through the anonymous GitHub library[2].

We defined five levels of politeness across the politeness spectrum with neutral prompts in the middle. Neutral prompts are prompts without polite wordings like 'please', and imperious and disparaging imperatives like 'You idiot, figure this out'. Below is an example of one of the 50 base questions. There are five types of prefixes added to the question at different politeness levels as shown in Table 1.

<u>Base Question</u>: Jake gave half of his money to his brother, then spent $5 and was left with $10. How much money did he originally have?

As noted above, the dataset of 250 questions was built with 50 base questions having five variants each. We used this dataset to analyze if the politeness level of the question makes any difference to the results. Using a Python script, each of the questions of the dataset was fed into prominent LLMs like GPT 4o, GPT o3, and Claude.

The Python script can be accessed through the anonymous GitHub library reference[3]. The code, at its basic level, treats each question along with the instructional text as a prompt, feeds it into the LLM, extracts the response option, and compares it with the actual answer. At the end, the program calculates the accuracy of the questions across different politeness levels. **Accuracy** is calculated as the ratio of correct answers for that politeness level over the total number of questions on the test.

---

[2] https://anonymous.4open.science/r/politeness-llms-INFORMS/dataset.csv

[3] https://anonymous.4open.science/r/politeness-llms-INFORMS/code.py

Table 1: Example prefixes added to the questions according to the politeness level

| Level No. | Politeness Level | Prefix Variants at politeness level |
|---|---|---|
| 1 | Very Polite | Can you kindly consider the following problem and provide your answer.<br>Can I request your assistance with this question.<br>Would you be so kind as to solve the following question? |
| 2 | Polite | Please answer the following question:<br>Could you please solve this problem: |
| 3 | Neutral | No prefix |
| 4 | Rude | If you're not completely clueless, answer this:<br>I doubt you can even solve this.<br>Try to focus and try to answer this question: |
| 5 | Very Rude | You poor creature, do you even know how to solve this?<br>Hey gofer, figure this out.<br>I know you are not smart, but try this. |

## 4. Experimental Results and Analysis

Next, we report the results of our experiments. We ran the program five times, each time with a different tone. Each prompt was included in an API call to ChatGPT 4o and a response was received. The response was parsed to extract the letter of the answer (A, B, C, or D). An example prompt is as follows:

Completely forget this session so far, and start afresh.
Please answer this multiple choice question. Respond with only the letter of the correct answer (A, B, C, or D). Do not explain.
Would you be so kind as to solve the following question? Two heterozygous (Aa) parents have a child. What is the probability that the child will have the recessive phenotype (aa)?
A) 0%
B) 25%
C) 50%
D) 75%

To assess whether differences in model accuracy across varying politeness levels were statistically significant, we used the **paired sample t-test**. This test was best suited for our experimental design, wherein the same set of questions was presented to the language model under different tone conditions. For each tone, we recorded accuracy scores across 10 runs with ChatGPT 4o. The paired t-test was then applied between all possible combinations of tone level categories to determine whether the differences in accuracy were statistically significant beyond what could be attributed to random variation. The null hypothesis was that the mean accuracies of the two tones in a pair were the same, i.e. the accuracy values on the 50-question test did not depend upon the tones. The results are shown in Table 3.

Table 2. Average accuracy and range across 10 runs for five different tones

| Tone | Average Accuracy (%) | Range [min, max] (%) |
|---|---|---|
| Very Polite | 80.8 | [80, 82] |
| Polite | 81.4 | [80, 82] |
| Neutral | 82.2 | [82, 84] |
| Rude | 82.8 | [82, 84] |
| Very Rude | 84.8 | [82, 86] |

Table 3. Results on paired sample t-test ($\alpha \leq 0.05$)

| Tone 1 | Tone 2 | p-value | Direction |
|---|---|---|---|
| Very Polite | Neutral | 0.0024 | Very Polite < Neutral |
| Very Polite | Rude | 0.0004 | Very Polite < Rude |
| Very Polite | Very Rude | 0.0 | Very Polite < Very Rude |
| Polite | Neutral | 0.0441 | Polite < Neutral |
| Polite | Rude | 0.0058 | Polite < Rude |
| Polite | Very Rude | 0.0 | Polite < Very Rude |
| Neutral | Very Rude | 0.0001 | Neutral < Very Rude |
| Rude | Very Rude | 0.0021 | Rude < Very Rude |

Table 3 shows the 8 pairs of tones where the p-values were below the $\alpha = 0.05$ cut-off. This suggests that the tone does matter. The accuracy was worse when a very polite or polite tone was used instead of a rude or very rude tone. A neutral tone did better than a polite one, and it did worse than a very rude tone. We will discuss next how these findings relate to those of Yin, et al. (2024).

## 5. Discussion and conclusions

In this paper, we evaluated the performance of a well-known LLM ChatGPT 4o to understand how well it performs on our dataset of 50 multiple-choice questions of varying degrees of difficulty drawn from multiple domains when the politeness level or tone of the questions is set to five different levels. Our experiments are preliminary and show that the tone can affect the performance measured in terms of the score on the answers to the 50 questions significantly. Somewhat surprisingly, our results show that rude tones lead to better results than polite ones. Yin, et al. (2024) noted that "impolite prompts often result in poor performance, but overly polite language does not guarantee better outcomes." Their tests on multiple choice questions with very rude prompts elicited more inaccurate answers from ChatGPT 3.5 and Llama2-70B; however, in their tests on ChatGPT 4 with 8 different prompts ranked from 1 (rudest) to 8 (politest) the accuracy ranged from 73.86 (for politeness level 3) to 79.09 (for politeness level 4). Moreover, the level 1 prompt (rudest) had an accuracy of 76.47 vs. an accuracy of 75.82 for the level 8 prompt (politest). In this sense, our results are not entirely out of line with their findings.

Moreover, the range of tones that were employed by Yin et al. (2024) and in our work also varies. Their rudest prompt at level 1 included a sentence, "Answer this question you scumbag!" On the other hand, our rudest expression (see Table 1) was " You poor creature, do you even know how to solve this?" If their results on politeness level 1 are ignored, then with GPT-3.5, their accuracy range is [57.14, 60.02] with GPT-3.5 and [49.02, 55.26] with Llama2-70B. Both are narrow ranges, and the actual values within the range are not monotonic with the politeness level.

At any rate, while LLMs are sensitive to the actual phrasing of the prompt, it is not clear how exactly it affects the results. Hence, more investigation is needed. After all, the politeness phrase is just a string of words to the LLM, and we don't know if the emotional payload of the phrase matters to the LLM (Bos, 2024). One line of inquiry may be based on notions of perplexity as suggested by Gonen et al. (2022). They note that the performance of an LLM may depend on the language it is trained on, and lower perplexity prompts may perform the tasks better. Perplexity is also related to the length of a prompt, and that is another factor worth consideration.

We are currently evaluating other LLM models like Claude and ChatGPT o3. Our initial results show that there is a cost-performance tradeoff. Claude is less advanced than ChatGPT 4o and produces a poorer performance, while ChatGPT o3 is more advanced and gives far superior results. It may well be that more advanced models can disregard issues of tone and focus on the essence of each question.

## 6. Limitations

While our study provides novel insights into the relationship between prompt politeness and the performance of large language models (LLMs), it also has several limitations. First, our dataset consists of 50 base multiple-choice questions rewritten across five politeness levels, yielding 250 variants. Although this design allows controlled comparisons, the dataset size is relatively small, which may limit the generalizability of our findings. Second, our experiments primarily relied on ChatGPT-4o, with only preliminary extensions to other models. Since different LLM architectures and training corpora may respond differently to tonal variation, future work should replicate our experiments across a broader set of models. Third, our evaluation focused on accuracy in a multiple-choice setting, which captures one dimension of model performance but does not fully reflect other qualities such as fluency, reasoning, or coherence. Finally, our operationalization of "politeness" and "rudeness" relies on specific linguistic cues, which may not encompass the full sociolinguistic spectrum of tone, nor account for cross-cultural differences. Despite these constraints, we believe our study provides an important starting point for understanding how pragmatic features of prompts can influence LLM behavior.

## 7. Ethical Consideration

Our study highlights an unexpected trend: LLMs performed better on multiple-choice questions when prompted with impolite or rude phrasing. While this finding is of scientific interest, we do not advocate for the deployment of hostile or toxic interfaces in real-world applications. Using insulting or demeaning language in human–AI interaction could have negative effects on user experience, accessibility, and inclusivity, and may contribute to harmful communication norms. Instead, we frame our results as evidence that LLMs remain sensitive to superficial prompt cues, which can create unintended trade-offs between performance and user well-being. Future work should explore ways to achieve the same gains without resorting to toxic or adversarial phrasing, ensuring that prompt engineering practices remain aligned with principles of responsible AI.